\documentclass[acmtog]{acm}
\usepackage{color}
\usepackage{subfigure}
\usepackage{units}
\usepackage{expl3}
\usepackage{amsmath}
\usepackage{multirow}
\usepackage{colortbl}
 \usepackage{tikz-cd}
 
\definecolor{Yellow}{rgb}{1,1, 0.6}
\definecolor{Red}{rgb}{1, 0.6, 0.6}
 
\usepackage{booktabs}
\setcitestyle{square}

\renewcommand\footnotetextcopyrightpermission[1]{}
\makeatletter
\renewcommand\@formatdoi[1]{\ignorespaces}
\makeatother

 \newcommand{\DeCurto} {\mathop {\rm c\&z} \nolimits}
 
\begin{document}

\title{\huge Doctor of Crosswise: Reducing Over-parametrization in Neural Networks}

\author{J. D. Curt\'o $^{*,1,2,3,4}$, I. C. Zarza $^{*,1,2,3,4}$, K. Kitani$^{2}$, I. King$^{1}$, and M. R. Lyu$^{1}$.}
\affiliation{\institution{\\$^{1}$The Chinese University of Hong Kong. $^{2}$Carnegie Mellon. \\$^{3}$Eidgen\"ossische Technische Hochschule Z\"urich. $^{4}$City University of Hong Kong.\\ $^{*}$Both authors contributed equally.}}

\renewcommand{\shortauthors}{Curt\'o, Zarza, Kitani, King and Lyu.}
\renewcommand{\shorttitle}{Doctor of Crosswise: Reducing Over-parametrization in Neural Networks.}

\authorsaddresses{\{curto,zarza,king,lyu\}@cse.cuhk.edu.hk, kkitani@cs.cmu.edu \\ \href{https://www.decurto.tw}{decurto.tw} \href{https://www.dezarza.tw}{dezarza.tw}}

\begin{teaserfigure}
\centering
\includegraphics[width=0.6\textwidth]{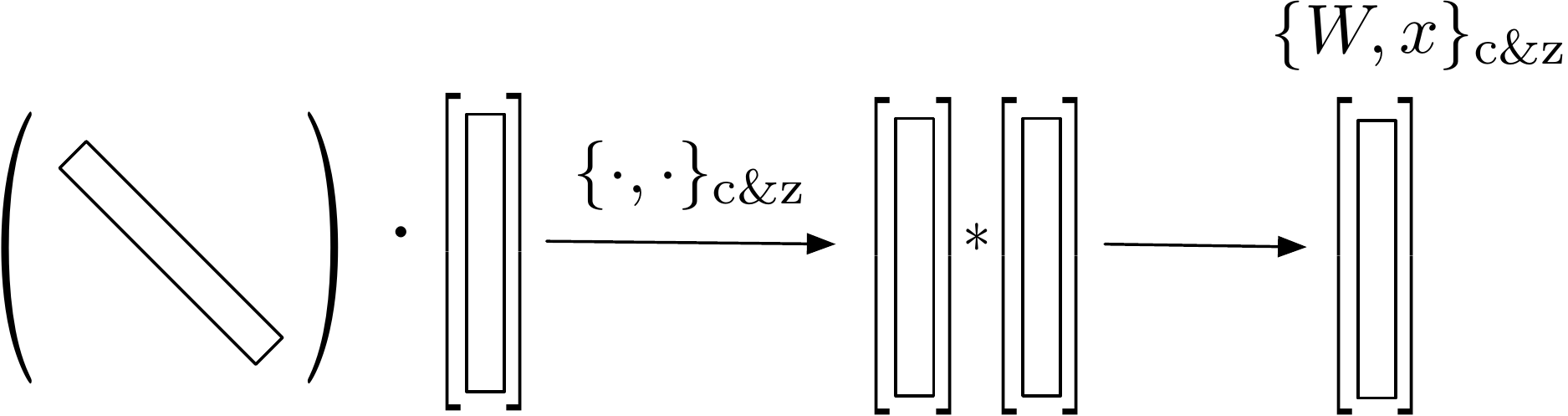}
   \caption{Dr. of Crosswise. Visual description of $\DeCurto$.}
\label{fgr:c_and_z}
\end{teaserfigure}

\ExplSyntaxOn
\newcommand\latinabbrev[1]{
  \peek_meaning:NTF . {
    #1\@}%
  { \peek_catcode:NTF a {
      #1.\@ }%
    {#1.\@}}}
\ExplSyntaxOff

\def\eg{e.g. }
\def\etal{et al. }


\begin{abstract}
Dr. of Crosswise proposes a new architecture to reduce over-parametrization in Neural Networks. It introduces an operand for rapid computation in the framework of Deep Learning that leverages learned weights. The formalism is described in detail providing both an accurate elucidation of the mechanics and the theoretical implications.
\end{abstract}

\maketitle
\fancyfoot{}
\thispagestyle{empty}


\section{Introduction}

Re-thinking how Deep Networks operate at large scale using current techniques is difficult. We build on the work in \cite{Curto17} to propose a fast variant named Dr. of Crosswise\footnote{Code is available at \href{https://www.github.com/curto2/dr_of_crosswise/}{https://www.github.com/curto2/dr\_of\_crosswise/}}. Our approach does not lose the ability to generalize while reduces vastly the number of parameters and improves training and testing speed. 
\\

Dr. of Crosswise substitutes common products matrix-vector in the architectures of Neural Networks by the use of simplified one-dimensional multiplication of tensors. Namely, it establishes a framework of learning where learned weight matrices are diagonal and optimized weights are considerably reduced. We introduce a new operation between vectors to allow for a fast computation.
\\

The structure of learning using matrices diagonal follows directly from the construction McKernel in \cite{Curto17} based on \cite{Le13, Rahimi08, Rahimi07}, which can be understood as a GAUSSIAN network \cite{Poggio17, Mhaskar17} of the form
\begin{align}
G(x) = \sum^{N}_{k=1} a_{k} \exp \left( - |x - x_{k}|^{2} \right), \qquad x \in \mathbb{R}^{d}.
\label{etn:gaussian}
\end{align}

We extend this idea to a more general framework, unconstrained in the sense that we no longer consider a form Gaussian, Equation \ref{etn:gaussian}, but any possible non-linearity (for instance ReLU). That is to say, Deep Learning.


\section{Deep Learning}

Successes in Neural Networks range across all domains, from Natural Language Processing \cite{Mikolov13,Pennington14,Devlin18} to Computer Vision \cite{Goodfellow14, Long14, Ren15,Simonyan15,Girshick15,Radford16,Yu16,Salimans16,He16,He17,Karras18,Curto17_2}, going through Automatic Structural Learning \cite{Cortes17} or Data Augmentation \cite{Cubuk18}. Techniques to improve aspects of current architectures have been widely explored \cite{Tremblay18, Acuna18, Coleman17}. Significant progress has also been made by combining Deep Learning for extraction of features with Reinforcement Learning \cite{Duan16, Levine16}.


\section{Doctor of Crosswise}
The way to go on Deep Learning entangles the following formalism

\begin{align}
y = \max(Wx+b,0)
\label{etn:formalism}
\end{align}

where matrix $W$ and vector $b$ are the weights and biases learned by assignment of credit \cite{Bengio93, Lecun98}, videlicet backpropagation.
\\

Dr. of Crosswise substitutes products matrix-vector by

\begin{align}
y = \max(\hat{W}x+b,0)
\label{etn:deeplearning}
\end{align}
where $\hat{W}$ is a matrix with form diagonal

\begin{align}
\hat{W} = \begin{bmatrix} 
    W_{1} & 0 & \dots & 0 \\
    0 & W_{2} & \ddots & \vdots \\
    \vdots & \ddots & \ddots & 0 \\
    0 &   \dots & 0     & W_{n}
    \end{bmatrix}.
\end{align} 

We can now factorize Equation \ref{etn:deeplearning} by the use of a new operand between vectors $\{\cdot, \cdot\}_{\DeCurto}$

\begin{align}
y = \max(\{c,x\}_{\DeCurto}+b,0)
\label{etn:crosswise}
\end{align}
where $z$ is a vector that holds the diagonal of $\hat{W}$, $c=[W_{1} \dots W_{N}]$.
\\

The operand defined by $\{\cdot, \cdot\}_{\DeCurto}$ is similar to a HADAMARD product and does the following operation. Given vectors $c$ and $x$ it computes
\begin{align}
\{c, x\}_{\DeCurto} =   \begin{bmatrix} c_{1}x_{1} \\ c_{2}x_{2} \\ \vdots \\ c_{n}x_{n}\end{bmatrix}
\end{align}
where $c$ holds the non-zero elements of a matrix diagonal.
\\

Note that Equation \ref{etn:crosswise} is equivalent to \ref{etn:deeplearning}. In other words, the new operator $\DeCurto$ helps factorize products between matrix diagonal and vector in form vector-vector. All the notation is preserved but for a change in the definition of the product. 
\\

We can understand this new operation between vectors as a product component-wise that scales each component by a given factor, Figure \ref{fgr:c_and_z}.
\\

Furthermore, the factors that need to be learned reduce from $n \times n$ to $n$ for each given layer, a momentous reduction of learned parameters and time of computation. Considering the compositionality of the problem, where architectures are build as a stack of many of these formalisms, Equation \ref{etn:formalism}, the improvements are considerably remarkable.
\\

\subsection{Extension to Higher-order}
\begin{figure}[H]
\centering
\includegraphics[scale=0.4]{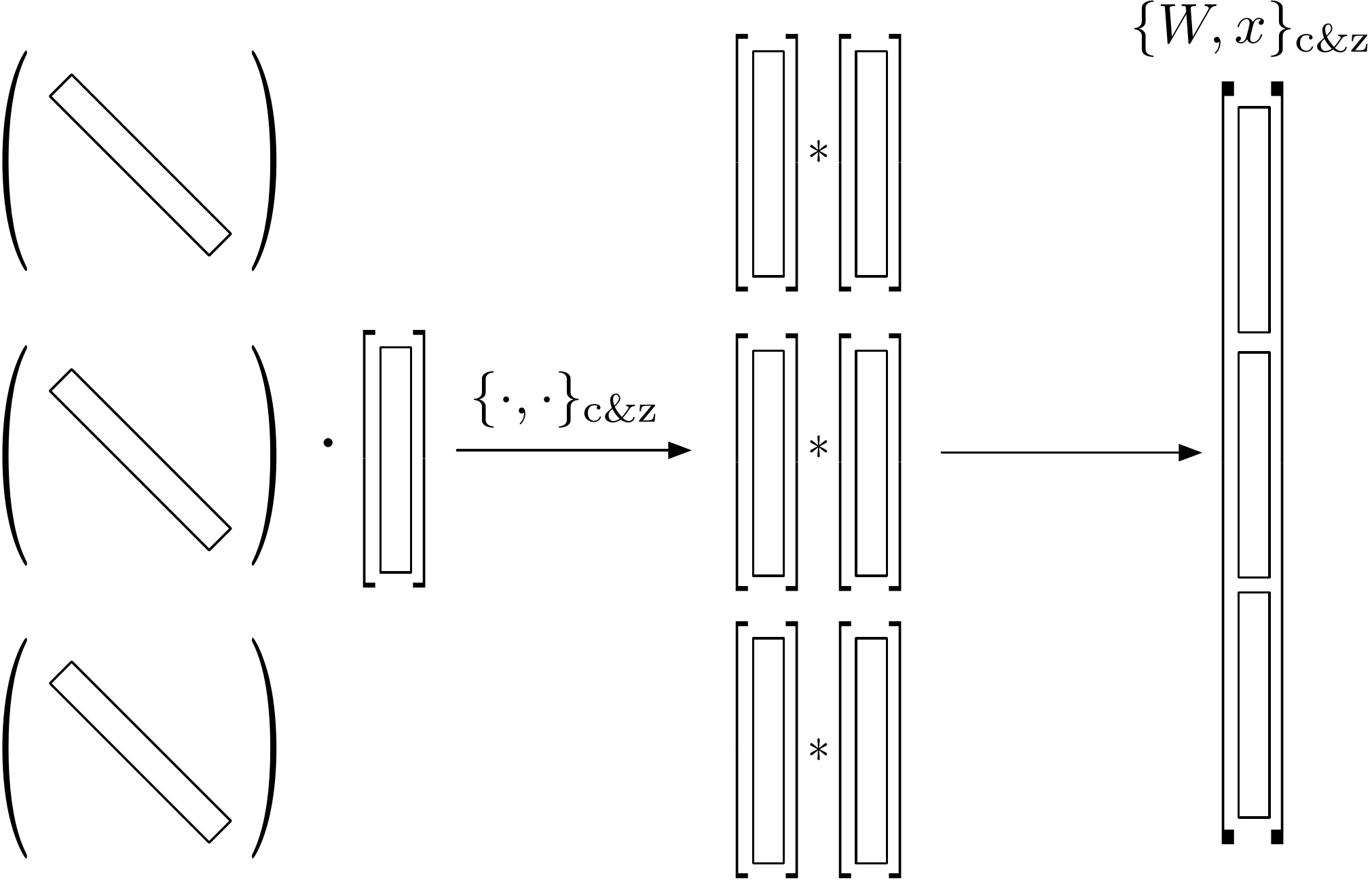}
\caption{Dr. of Crosswise. Visual description of $\DeCurto$.}
\label{fgr:dr_of_crosswise}
\end{figure}

If we now consider the setting where we have multiple input data and the need to expand it to higher-dimensional spaces from layer to layer, as it is normally done in Deep Learning. We have to consider several facts. Given a matrix $W$ of dimension $M \times N$ and input vector $x$ of dimension $N$ we will generate an output vector with size proportional to $N$. $W$ will be substituted by a set of matrices diagonal whose cardinal will be the closest multiplicity of $N$ to $M$, see Figure \ref{fgr:dr_of_crosswise}. In this way, we will need a product that operates element-wise between the elements of each matrix diagonal and the input vector. Considering a varying number of input vectors, this mathematical operation is very close to a KHATRI RAO product, which is the matching columnwise KRONECKER product \cite{Kolda09}.

These products between matrices have useful properties:
\begin{align}
&(A \otimes B) (C \otimes D) = AC \otimes BD\\
&(A \otimes B)^{\dagger} = A^{\dagger} \otimes B^{\dagger}\\
&A \odot B \odot C = (A \odot B) \odot C = A \odot (B \odot C) \\
&(A \odot B)^{T}(A \odot B) = A^{T}A * B^{T}B \\
&(A \odot B)^{\dagger} = ((A^{T} A) * (B^{T}B))^{\dagger} (A \odot B)^{T}
\end{align}

where $\odot$ denotes KHATRI RAO product, $\otimes$ specifies KRONECKER product and $*$ means HADAMARD product. $A\dagger$ is the MOORE PENROSE pseudo-inverse of $A$. 
\\
  
Take special note on the fact that for example, to transform an input vector of size $4$, to an output vector of size $8$, in the standard formulation of Neural Networks we have to learn $32$ weights. If we consider Dr. of Crosswise, the number of learned parameters to do exactly the same operation reduces to $8$. More importantly, the number of multiplications and additions also drastically lowers.


\section{Rationale}

We start with an exordium on Kernel Methods \cite{Cortes95,Vapnik18}. Let $X$ be a measure space with $k: L^{2}(X \times X) \to \mathbb{R}$. We name $k$ a kernel if, and only if, there is some feature map $\phi: X \to \mathbb{H}$ into a separable HILBERT space $\mathbb{H}$ such that

\begin{align}
k(x,x') = \langle \phi(x),\phi(x') \rangle_{\mathbb{H}}.
\end{align}

In other words, $k$ is a kernel if, and only if, for some space $\mathbb{H}$ and map $\phi$ the following diagram commutes:

\begin{center}
\begin{tikzpicture}[commutative diagrams/every diagram]
\matrix[matrix of math nodes, name=m, commutative diagrams/every cell] {
X \times X \\
\mathbb{H} \times \mathbb{H} & \mathbb{R}. \\};
\path[commutative diagrams/.cd, every arrow, every label]
(m-1-1) edge node[swap] {$\phi$} (m-2-1)
(m-1-1) edge node {$k$}(m-2-2)
(m-2-1) edge node[swap]  {$\langle \cdot,\cdot \rangle_{\mathbb{H}}$}(m-2-2);
\end{tikzpicture}
\end{center}
Random Kitchen Sinks \cite{Rahimi07}
approximate this mapping of features $\phi$ by a FOURIER expansion in the
case of RBF kernels
\begin{align}
    k(x,x') = \int \exp{(i \langle w,x\rangle)} \exp{(-i \langle w,x'\rangle)} d\rho(w).
\end{align}
\cite{Le13, Yang14} present a fast approximation of the matrix $W$. Recent works on the matter \cite{Wu16, Yang15, Moczulski16, Hong17} use this technique to extract meaningful features.
\\

Dr. of Crosswise is motivated by the construction McKernel in \cite{Curto17}, where Deep Learning and Kernel Methods are unified by extending the use of the approximation of $W$, $\hat{Z}$, to a SGD optimization setting
\begin{align}
  \hat{Z} := \frac{1}{\sigma \sqrt{n}} C H G \Pi H B.
  \label{etn:mckernel}
\end{align}
Here $C, G$ and $B$ are matrices diagonal, $\Pi$ is a random
matrix of permutation and $H$ is the Hadamard. Whenever the number
of rows in $W$ exceeds the dimensionality of the data, simply
generate multiple instances of $\hat{Z}$, drawn i.i.d., until the required
number of dimensions is obtained. 
\\

The key idea behind it is that the FOURIER transform diagonalizes the integral operator.
\\

This can be best seen as follows. Considering the operator of convolution working on the complex exponential $f(z)=\exp(ikz)$
\begin{align}
g(z) = (f*r)(z)= \int_{-\infty}^{\infty} f(\tau)r(z-\tau) d\tau.
\end{align}
Then
\begin{align}
g(z) = \lambda f(z)
\end{align}
where $\lambda = R(k)$ and $R$ is the FOURIER transform of $r$.
\\

We build on these ideas to pioneer a framework of Deep Learning where the formalism used entangles matrices diagonal.


\bibliographystyle{ACM-Reference-Format}

\end{document}